\documentclass[letterpaper]{article}
\usepackage{aaai}
\usepackage{times}
\usepackage{helvet}
\usepackage{courier}
\usepackage{graphicx}
\usepackage{mathrsfs}
\usepackage{color}
\usepackage{listings}
\usepackage{caption}
\usepackage{subcaption}
\usepackage{hyperref}
\usepackage{multirow}
\usepackage{amsmath,amsthm,amssymb}
\usepackage{subcaption}
\usepackage[utf8]{inputenc}

\begin{document}
\title{CFR-p $\colon$ Counterfactual Regret Minimization with Hierarchical Policy Abstraction, and its Application to Two-player Mahjong}
\author{Shiheng Wang  \\
pkuwsh@gmail.com  } 
\maketitle

\begin{abstract}
\begin{quote}
Counterfactual Regret Minimization(CFR) has shown its success in Texas Hold’em poker. We apply this algorithm to another popular incomplete information game, Mahjong. Compared to the poker game, Mahjong is much more complex with many variants. We study two-player Mahjong by conducting game theoretical analysis and making a hierarchical abstraction to CFR based on winning policies. This framework can be generalized to other imperfect information games.
\end{quote}
\end{abstract}

\section{Introduction to CFR}\label{sec:cfr}
\subsection{Normal-form and Extensive-form Games}
    A finite normal-form game is a tuple $(N, A, u)$, where
    \begin{itemize}
      \item $N=\{1,...,n\}$ is a finite set of n players.
      \item $A = S_1 \times ... \times S_n$ is the set of all action profiles, where $S_i$ is a finite set of actions available to player $i$. Each vector $a \in A$ is called an action profile (or outcome).
      \item $u = (u_1,...,u_n)$, where $u_i : A \mapsto \mathbb{R}$ is a real-valued utility (also called payoff or reward) function for player i.
    \end{itemize}

    A normal-form game is zero-sum if the utilities of all players sum up to be zero for any given outcome, i.e. $\forall a\in A, \sum_{i\in N} u_i(a)=0$. In constant-sum games, the utilities sum up to a constant value. Constant-sum games can be reformulated into zero-sum games.

    A pure strategy picks one action with probability 1, while a mixed strategy assigns a distribution over actions, denoted with $\sigma$.
    $\sigma_i(s)$ refers to the probability that player $i\in N$ picks action $s \in S_i$. $-i$ refers to player $i$'s opponents. The expected utility of player $i$ can be computed as,
    $$ u_i(\sigma_i, \sigma_{-i}) = \sum_{s\in S_i} \sum_{s' \in S_{-i}} \sigma_i(s) \sigma_{-i}(s') u_i(s, s').$$
    Given all other players' strategies, a best response for player $i$ maximizes its expected utility. When every player is playing a best response to other players' strategies, the combination of strategies is called a Nash Equilibrium, where no player can get higher utility by deviation.

    In a normal-form game, players take actions simultaneously. In a sequential game, however, a play consists of a sequence of actions. A sequential game is formed by a extensive-form game, where the game tree is formed of states with edges transiting from state to state.

    A state can be a chance node or decision node.
    A chance node assigns the outcome of a chance event, so each edge corresponds to an outcome with its probability.
    At a decision node, the edges represent the actions and their related successor states.
    Each decision node in the game tree is contained within an information set, which contains one active player and all information available to him. One information set may contain more than one game state.

    Let $A$ denote the set of all game actions, $I$ denote an information set, and $A(I)$ denote the set of legal actions for information set $I$. Let $t$ and $T$ denote time steps. A strategy $\sigma_i^t$ maps player i's information set $I_i$ to $\Delta(A(I_i))$, that is, a probability distribution over legal actions. At time $T$, all players' strategies form a strategy profile $\sigma^t$, and a strategy profile excluding player i is denoted as $\sigma_{-i}$. Let $\sigma_{I\to a}$ denote a profile equivalent to $\sigma$, except that action $a$ is always chosen at information set $I$.

    A history $h$ is a sequence of actions (including chance outcomes) starting from the game root. Let $\pi^\sigma(h)$ be the reach probability of the game history $h$ with strategy profile $\sigma$. Similarly, let $\pi^\sigma(I)$ be the probability of reaching information set $I$ through all possible histories in $I$, i.e. $\pi^\sigma(I) = \sum_{h\in I} \pi^\sigma(h)$.

    All extensive-form games have an equivalent normal-form representation.

\subsection{Regret Minimization}
    This introduction follows from \cite{neller2013introduction}.

    {\it Regret Matching}, introduced by Hart and Mas-Colell in 2000, has sparked a revolution in computer game play of some of the most difficult incomplete-information games. Players reach equilibrium by tracking regrets for past plays, making future plays proportional to positive regrets.

    Regret of not having chosen an action is defined as the difference between the utility of that action and the utility of the action that is actually chosen, with respect to the fixed choices of other players. Formally, for action profile $a \in A$, let $s_i$ be player $i$'s action and $s_{-i}$ be the actions of all other players. Suppose $s_i'$ is substituted for $s_i$, then after the play, player $i$'s regret for not have played $s_i'$ is $u(s_i', s_{-i})- u(s_i, s_{-i}) $.

    In order to chose the action that has the largest regret and meanwhile not totally predictable and thus exploitable, one player may take regret matching, where actions are selected at random with a distribution that is proportional to positive regrets.

    When the game is repeated for a number of rounds, regrets of previous rounds accumulate. 
    Over time, for two-player zero-sum games, regret matching converges to a correlated equilibrium.
    
    Our tutorial demo for regret matching is available at, 
    \begin{center}
        \url{https://github.com/workplay/CFR}
    \end{center}
    
\subsection{Counterfactual Regret Minimization}
    Regret matching is only applicable to normal-form games, while CFR works for extensive-form games.
    This introduction also follows from~\cite{neller2013introduction}, and
    the details can be found in~\cite{lanctot2009monte} and~\cite{zinkevich2008regret}.

    The counterfactual reach probability of information set $I$, $\pi_{-i}^{\sigma}(I)$, is the probability of reaching $I$ with strategy profile $\sigma$ except that, we treat current player $i$'s actions to reach the state as having probability 1. ``Counterfactual'' here means that player $i$'s strategy is modified to have intentionally played to information set $I_i$. In other words, the probabilities that factually came into player $i$'s play is excluded from the computation.

    Let $Z$ denote the set of all terminal game histories (from root to leaf). Then proper prefix $h\sqsubset z$ for $z\in Z$ is a nonterminal game history. $u_i(z)$ is player $i$'s utility of terminal history $z$. Define the counterfactual value at nonterminal history $h$ as:
    \begin{equation}\label{eq:cfv}
    v_i(\sigma, h) = \sum_{z\in Z, h\sqsubset z} \pi_{-i}^{\sigma}(h)\pi^{\sigma}(h,z)u_i(z)
    \end{equation}
    The counterfactual regret of not taking action $a$ at history $h$ is defined as:
    \begin{equation}\label{eq:cfrh}
        r(h,a)=v_i(\sigma_{I\to a}, h) - v_i(\sigma, h)
    \end{equation}
    The counterfactual regret of not taking action $a$ at information set $I$ is:
    \begin{equation}\label{cfrI}
        r(I,a) = \sum_{h\in I} r(h,a)
    \end{equation}
    Let $r_i^t(I,a)$ refer to the regret when players use $\sigma^t$ of not taking action $a$ at information set $I$ belonging to player $i$. The cumulative counterfactual regret is defined as:
    \begin{equation}\label{cfr_culmulative}
        R_i^T(I,a) = \sum_{t=1}^{T} r_i^t(I,a)
    \end{equation}

    The regret of action $a$ is the difference between the value of always choosing action $a$ and the expected value of strategy $\sigma$, weighted by the probability that other players (including chance player) will play to reach the node.

    Suppose the nonnegative counterfactual regret is defined as $R_i^{T,+}=\max (R_i^T(I,a), 0)$, then the new strategy can be obtained from cumulative regrets.
    \\If $\sum_{a\in A(I)} R_i^{T,+}(I,a)>0$,
    \begin{equation}\label{eq:cfr_strategy}
    \sigma_i^{T+1}(I,a) = \frac{R_i^{T,+}(I,a)}{\sum_{a\in A(I)} R_i^{T,+}(I,a)}.
    \end{equation}
    Otherwise take a random strategy,
    \begin{equation}
    \sigma_i^{T+1}(I,a) = {1} \big / {| A(I) | }    
    \end{equation}

    For each information set, equation~(\ref{eq:cfr_strategy}) computes a strategy of which action probabilities are proportional to the positive cumulative regrets. CFR computes the utility of each action recursively. Regrets are computed from returned values, and the values of the current node is finally computed and returned. The average strategy profile at information set $I$ approaches an equilibrium as $T\to \infty$.

\section{Two-Player Mahjong}
\subsection{Rules} \label{sec:rules}

\begin{table}[htbp]
\centering
\caption{Mahjong Tile Count}
\label{tab:tilecount}
\begin{tabular}{|c|c|c|}
\hline
\textbf{Category}        & \textbf{Name}  & \textbf{Count} \\ \hline
\multirow{3}{*}{Simples} & Dots       & 36    \\ \cline{2-3}
                         & Bamboo     & 36    \\ \cline{2-3}
                         & Characters & 36    \\ \hline
\multirow{2}{*}{Honors}  & Winds      & 16    \\ \cline{2-3}
                         & Dragons    & 12    \\ \hline
\multirow{2}{*}{Bonus}   & Flowers    & 4     \\ \cline{2-3}
                         & Seasons    & 4     \\ \hline
\multicolumn{2}{|c|}{Total}           & 144   \\ \hline
\end{tabular}
\end{table}

\begin{figure*}[htbp]
  \centering
  \includegraphics[width=0.85\textwidth]{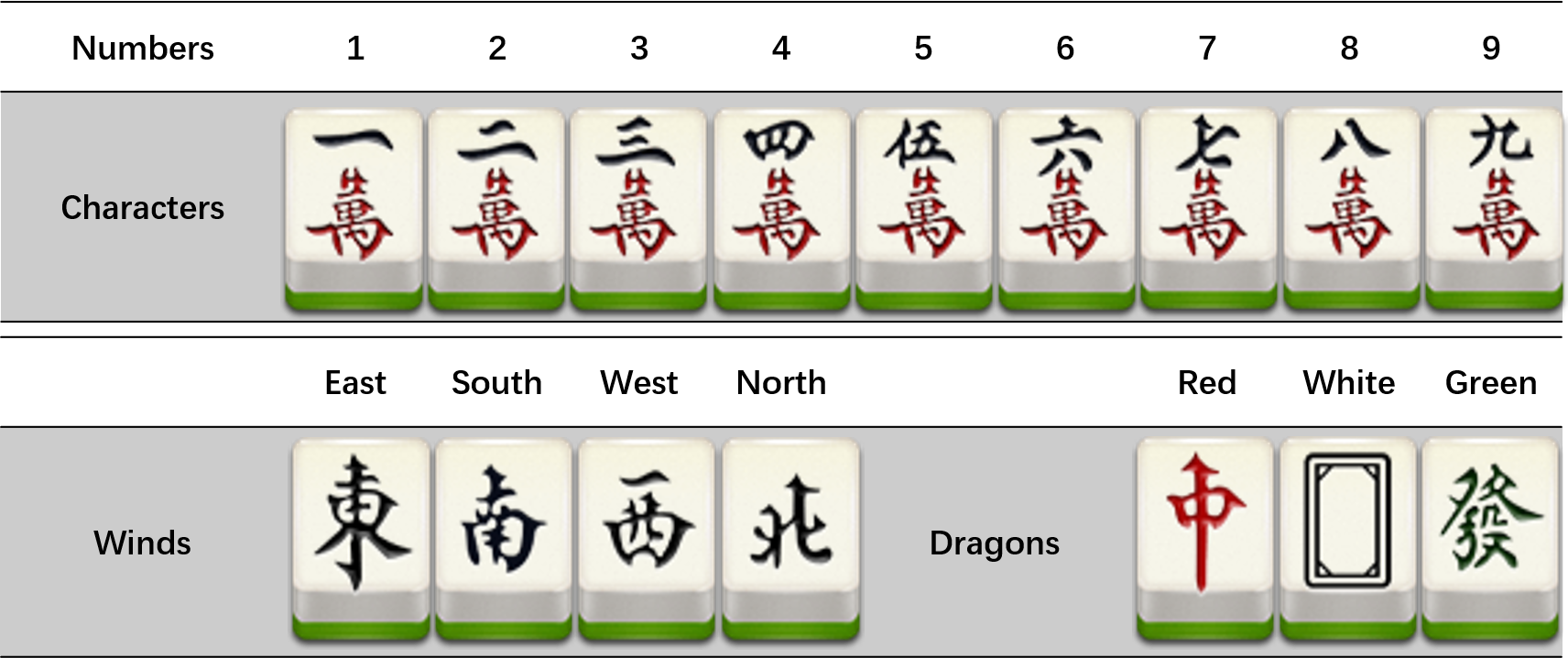}
  \caption{Two-player Mahjong Tiles}\label{fig:tiles}
\end{figure*}

    Mahjong is a tile-based game developed in China, which is commonly played by four players. There is a detailed elaboration on Wikipedia\footnote{\url{https://en.wikipedia.org/wiki/Mahjong}}, which is abbreviated to this concise instruction. As is shown in Table~\ref{tab:tilecount}, a typical set of Mahjong tiles usually has at least 136 tiles.
    There are 3 suits of simples and in each suit the tiles are numbered from 1 to 9. The suits are bamboos, dots, and characters. There are 4 identical copies of each simples tile totaling 108 simples tiles.
    There are two different sets of Honors tiles: Winds and Dragons. The Winds are East, South, West, and North. The Dragons are Red, Green, and White. These tiles have no numerical sequence and there are four identical copies of each Honors tile, for a total of 28 Honors tiles. There are two sets of Bonus tiles: Flowers and Seasons. When drawn, the Bonus tile is not added into a player's hand but are instead set aside for scoring purposes, and an extra tile is drawn in replacement of the Bonus tile.

    Two-player Mahjong is a simplified variant designed by Tencent
    \footnote{\url{https://majiang.qq.com/index.html}},
    which is played with a set of 68 tiles. All Dots and Bamboos are excluded, and there are only Characters, Honors and Bonus. As Bonus don't affect the rules of the game, they are not considered in our analysis.
    The tiles are displayed in Figure~\ref{fig:tiles}.

    Although there are fewer tiles and players, the basic rules remain the same. Each player begins by receiving 13 tiles, and in turn players draw and discard tiles until they complete a legal hand using the 14th drawn tile to form 4 melds (or sets) and an eye (two identical tiles).
    Melds are groups of tiles within the player's hand, consisting of either a Pong (three identical tiles), a Kong (four identical tiles), a Chow (three Simple tiles all of the same suit, in numerical sequence). Whenever a Kong is formed, that player must draw an extra tile from the end of the wall(face down tiles on the table) and then discard a tile. Melds may be formed by drawing a tile from the wall, or by seizing another player's discard.
    A player can also win with other special hands, like seven Eyes.

    Points are obtained by matching the winning hand with different values. For simplicity, we only consider most popular patterns of winning. PongPongHu(4 Pongs and 1 Eye) or QiDui(7 Eyes) gets two points, and all other winning patterns get one point.
    We choose two-player Mahjong to study algorithms for incomplete-information games. This framework can be generalized to other variants in the future.

\subsection{Extensive-Form Game Representation}
    Each player in turn, in counterclockwise direction, draws a tile from the wall; then this player proceeds to discard a tile. The discarded tile is thrown into the centre and the other players have an opportunity to seize the discarded tile; if no one takes it, the turn continues to the next player. Play continues this way until one player has a legal winning hand and calls out "Hu" while revealing their hand.

    \begin{figure*}[t]
      \centering
      \includegraphics[width=0.9\textwidth]{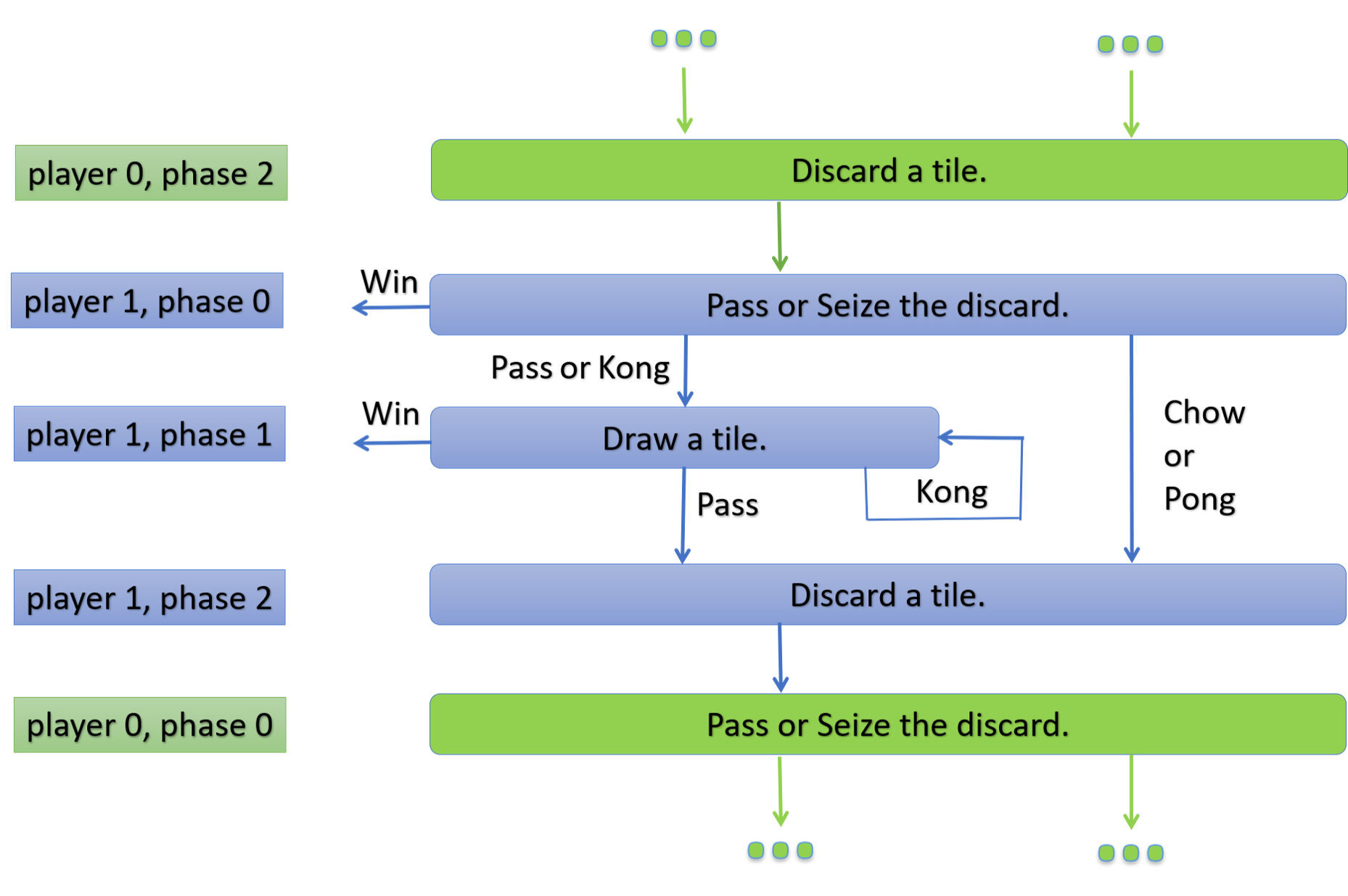}
      \caption{Extensive-Form Game Representation}\label{fig:extensive-form}
    \end{figure*}
    As the normal counterclockwise order may be interrupted, a neat extensive-form game~\cite{shoham2008multiagent} representation is necessary in order to conduct game theoretical analysis as well as to implement algorithms. There are two players 0 and 1, and each player's turn is separated into three phases 0, 1 and 2. Figure~\ref{fig:extensive-form} shows one turn for player 1. In phase 1, although there is an edge pointing from node ``Draw a tile.'' to itself by action Kong, the player actually draws a tile before he acts Kong, thus the game goes to a new state, which doesn't result in a cycle in the game tree.

    \begin{itemize}
      \item In phase 0, the player can seize the discard by forming Chow, Pong, Kong, Win with tiles in his hand, otherwise he can only Pass. For Chow and Pong, he directly goes to phase 2, otherwise to phase 1.
      \item In phase 1, the player draws a tile from the wall. Possible legal actions are declaring Win, Kong and Pass. By choosing Kong, he reenters phase 1, while Pass leads to phase 2.
      \item In phase 2, the only legal action is to discard a tile from his hand.
    \end{itemize}

    \begin{table}[htbp]
      \centering
      \caption{Possible Legal Actions}\label{tab:actions}
      \begin{tabular}{|c|c|}
        \hline
        \textbf{Phase} & \textbf{Possible Legal Actions} \\ \hline
        0 & Chow, Pong, Kong, Win, Pass. \\ \hline
        1 & Kong, Win, Pass. \\ \hline
        2 & Discard a tile. \\
        \hline
      \end{tabular}
    \end{table}
    Possible legal actions are summarized in Table~\ref{tab:actions}. Notice that Chow, Pong, Kong, Win are legal actions only when certain patterns can be formed with other tiles in hand.

\subsection{Complexity}
    Suppose all hidden information are given, the complexity of two-player Mahjong is similar to Chess. The number of reachable positions on a chess board is estimated to be fewer than $10^{46}$~\cite{chinchalkar1996upper} .   As most actions in phase 0 and 1 are legal only with certain tiles in hand, we assume that both players take the action Pass in order to estimate the complexity of the game tree. In phase 3, however, during each round there are 14 tiles in hand, and the player chooses one of them to discard. Two-player Mahjong consists of 36 Character tiles and 28 Honor tiles, of which 26 tiles will be allocated to both players in the beginning. Therefore the game tree has more than $14^{38}$ leaves, and the complexity can reach $10^{43}$. Note that only one action is considered in phase 0 and 1, thus the complexity serves as a lower bound.

    As long as incomplete information is introduced, the game becomes much more complicated. Each permutation corresponds to a different game, and only discarded tiles are public information to both players.  There are 4 identical copies of 16 different tiles, and the number of permutations after shuffling can be calculated as,
    \begin{equation}\label{eq:permutations}
        \frac{A_{64}^{64}}{ (A_{4}^{4})^{16}} \approx 1.2 * 10^{22}.
    \end{equation}
    In total, the complexity reaches $10^{64}$. In comparison, two-player Limit Texas Hold'em poker only has $10^{18}$ leaves~\cite{sandholm2010state}.

    Mahjong is much more complex than Texas Hold’em poker, which is popular among literatures about incomplete information games. The most significant difference is that the hidden information keeps changing during the entire play. In Texas Hold’em poker, play begins with each player being dealt two cards face down, and this hidden information remains the same during the entire play. In Mahjong, however, hidden tiles in hand keep changing after every round. As a result, existing algorithms such as decomposition~\cite{burch2014solving} or subgame solving~\cite{brown2017safe} cannot be directly applied to Mahjong. Secondly, the amount of hidden information is much larger. There are only 2 face down cards in poker, but there are 13 hand tiles in Mahjong. Finally, each poker player only bets for at most five rounds, while in Mahjong, the number of rounds can be as large as 38.

    Note that two-player Mahjong is the most simplified version. Usually a Mahjong game is played by four players with 136 tiles.

\section{Hierarchical Policy Abstraction}\label{sec:abstraction}
    Because of the complexity caused by a lot of private hidden information that keeps changing, we cannot implement CFR on two-player Mahjong directly. We make many abstractions in order to reduce the complexity.

    \begin{figure}[t]
      \centering
      \includegraphics[width=\linewidth]{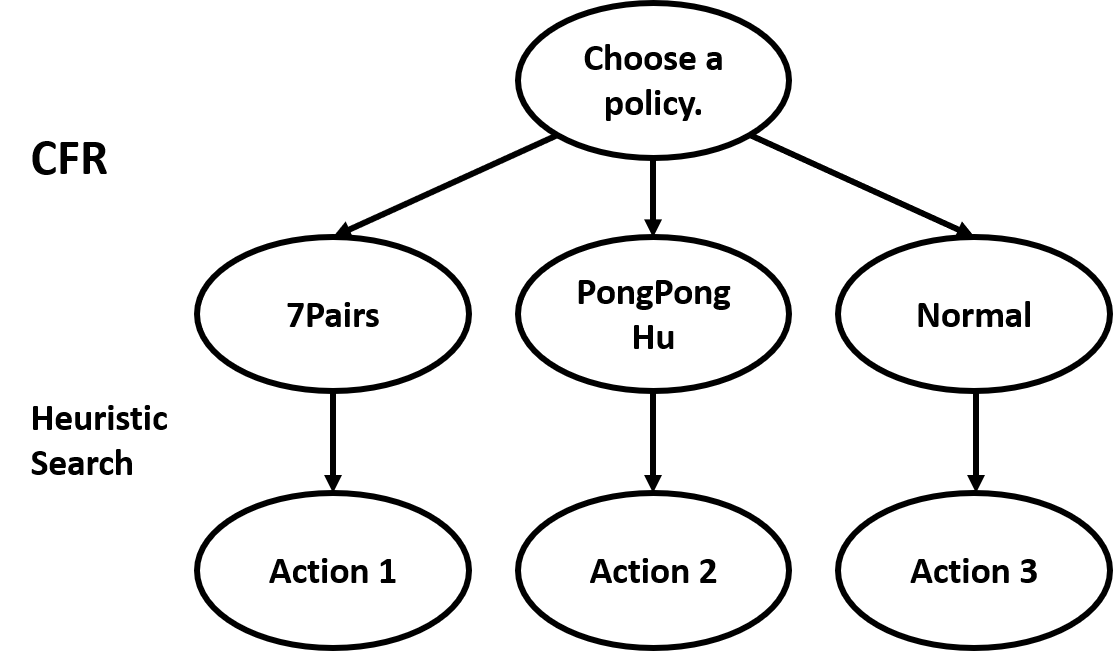}
      \caption{Hierarchical Policy Abstraction}\label{fig:policyAbstract}
    \end{figure}

    At high level, we conduct hierarchical policy abstraction. As is introduced in section~\ref{sec:rules}, there are several patterns of winning, such as QiDui, PongPongHu, etc. Based on the fact that there are several distinct patterns, and that winning with the same pattern usually receives the same points, we separate the task of the game into two parts:

    \begin{enumerate}
      \item Which winning pattern should we achieve?
      \item What is the most efficient way to achieve it?
    \end{enumerate}

    Suppose the complexity of both parts are $\mathcal{O}(T_1)$ and $\mathcal{O}(T_2)$. Compared to an end-to-end algorithm, the hierarchical policy abstraction significantly reduces the complexity from $\mathcal{O}(T_1*T_2)$ to $\mathcal{O}(T_1+T_2)$. In addition, at each node of the game tree, the number of legal actions plunges to the number of winning patterns. For instance, in two-player Mahjong, phase 3, the agent need to discard a tile out of 14 tiles in his hand, resulting in 14 possible legal actions. With policy abstraction, however, the number of legal actions becomes 3, i.e. to win with PongPongHu, to win with QiDui, or to win Normally without special patterns. Finally the number of leaves in the game tree can be as few as $3^38 \approx 1.35 \times 10^{18}$. Moreover, it's not necessary to change the policy at every round, especially when the agent is reaching the end of the game, so that the size of the game tree can be further reduced.

    Each separate sub-task can be solved independently by adopting an appropriate algorithm, such as rule-based search, supervised learning with professional players' game history, or reinforcement learning like CFR. For the 1st sub-task, which winning pattern can win most points should be concluded based on the statistics from self-play. As for the 2nd sub-task, since there is a clear indicator to the distance between the tiles in hand and the target winning pattern, rule-based search can help solve this problem, such as keeping melds in hand and dropping other single tiles.

     \begin{figure*}[t]
      \centering
      \includegraphics[width=0.9\textwidth]{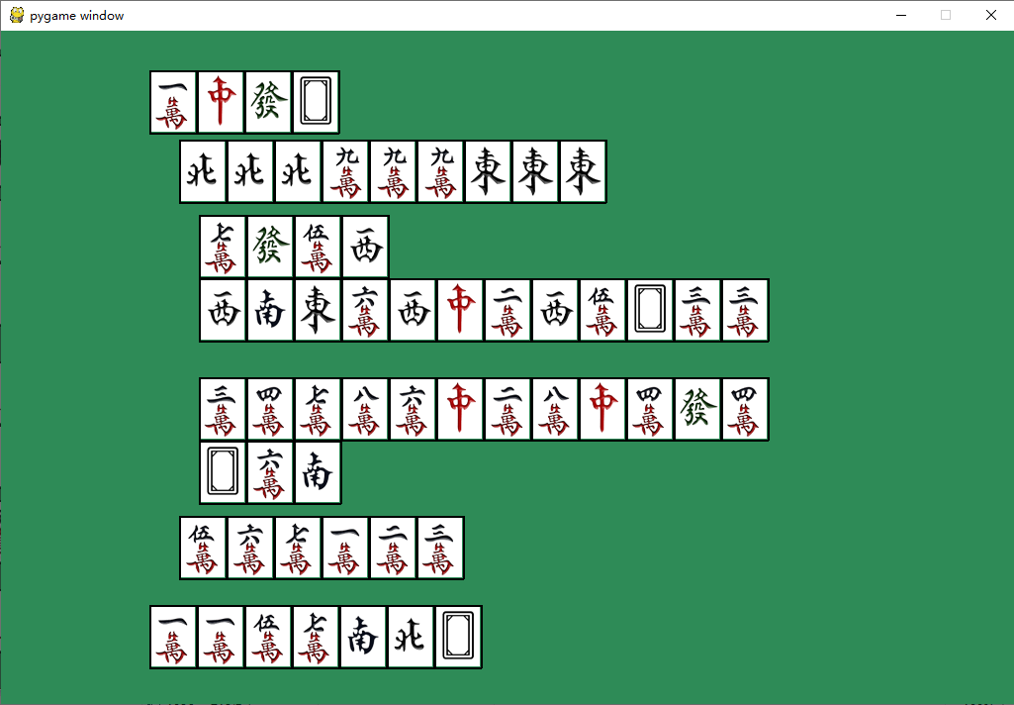}
      \caption{GUI for Testing}\label{fig:GUI}
    \end{figure*}

\section{Implementation}
     Our two-player Mahjong framework consists of three parts: the game logic, the AI and the GUI(graphical user interface) for testing.

\subsection{The Game Logic and the GUI for Testing}
     The game logic is designed following the extensive-form representation in Figure~\ref{fig:extensive-form}. It tracks tiles in hand as well as all hidden and dropped tiles on the table, determines legal actions in every phase, and writes down all game states to a log file.
     Some legal actions can be decided directly according to special patterns and tiles in hand, for example, whether hand tiles can form a pattern of Chow, Pong or Kong together with the tile dropped by the other player. Some other legal actions such as winning are much more complex, which requires the game logic to check whether the tiles in hand form any of the winning patterns.

     In purpose of supporting CFR and other search based algorithms, the game logic is designed recursively. At each node, it makes a copy of current game state, tries some legal actions, and returns the utility recursively from a leaf node. Since there is only one searching path from root to current game node each time, only linear memory is required while running search algorithms.

     The game logic supports both training and testing. In the training mode, it repeatedly traverse the game tree recursively, keeping a record of utilities for each action at each node. The choice of actions and recorded utilities is determined by the AI algorithm. When testing, however, it runs one possible path from root to leaf, records all game states and actions and writes them to a log file, which can be analyzed by GUI for testing.

     As is shown in Figure~\ref{fig:GUI}, the GUI for testing reads the log file and displays the entire game process graphically. It keeps track of all tiles in hand, all dropped tiles and the actions like Chow, Pong, Kong. Although only partial information can be observed by the agent, all hidden information is released to the tester. Whoever with some domain knowledge of this game can evaluate whether the AI has made a rational move or not.

\subsection{The AI}
    As is discussed in section~\ref{sec:abstraction}, policies are abstracted into actions according to winning patterns. For all three phases at every state there are only three actions that are abstracted from policies, i.e. PongPongHu, 7Pairs and Normal. There may be other policies in other variants of Mahjong, such as Defense, Same Color, etc. Defense, for example, decrease other players' probability of winning regardless his own hand tiles. Such policy is not suitable for two-player Mahjong because the game is zero-sum, thus defense also prevents the agent itself from winning.

    Each policy is realized by heuristic search, whose heuristic function considers the number of potential winning tiles, the number of melds and pairs, as well as the possible missing tiles which can form melds or pairs with hand tiles. In general, the search algorithms aim at maximize the probability of winning, ignoring winning points. The winning points, however, is considered by the higher level CFR algorithm that picks the appropriate winning pattern.

    \begin{table}
      \centering
      \begin{tabular}{|c|c|c|}
        \hline
        Bit & Meaning & Range \\ \hline
        0-5 & The serial number of round & 0 - 38 \\ \hline
        6-8 & The number of Pairs & 0 - 6 \\ \hline
        9-11 & The number of Pongs or Kongs & 0-4 \\ \hline
        12-15 & The number of Character tiles & 0-14 \\ \hline
        16-19 & The number of Wind tiles & 0-14 \\ \hline
      \end{tabular}
      \caption{Encoding Features}\label{tab:encode}
    \end{table}

    The CFR algorithm is to find which abstract policy can win most points in expectation at every decision node. The algorithm mainly follows from the description in section~\ref{sec:cfr}, which assigns a mixed strategy to every information set.  We only do CFR three times for each player in the beginning, middle and end of the game since it's not necessary to change policies in every round.  The accumulated regrets and strategies are saved into a file with a tabular representation, where each decision node is labeled by a compact encoding.

    Figure~\ref{tab:encode} shows how the labels of nodes are encoded by a single integer. Because of the huge amount of possible hand tiles and table tiles, we extract features of hand tiles and cluster information sets according to these features. Recall that there are basically three abstract actions, i.e. Normal, PongPongHu and 7Pairs. If there exists an explicit Chow in the hand, Normal is the only legal action so that the corresponding decision node will not be saved.

    While playing a game, the AI loads the node labeled by the same information set as current hand tiles, then it picks one legal abstract policy based on the accumulated regrets. The real action will be generated automatically by the underlying heuristic search algorithms.

\section{Evaluation}
    The notations in this section follow from~\cite{johanson2011accelerating}.

    To evaluate an extensive game strategy, there are typically two options. The first one is to organize a tournament consisting of several strategies, like the Annual Computer Poker Competition. The second one is to compute the worst-case performance of a strategy. It is infeasible for large scale games because it requires to traverse the entire game tree.
    We choose to compute the worst-case performance of the abstracted game in this paper, and will organize online tournaments with human players in the future.

    A strategy for player i, $\sigma_i \in \Sigma_i$, is a function that assigns a probability distribution over actions to each information set $I$. A strategy profile, $\sigma \in \Sigma$, consists of a strategy for each player. $\sigma_{-i}$ refers to all strategies in $\sigma$ except for $\sigma_i$. $u_i(\sigma)$ is the utility for player $i$ under strategy profile $\sigma$.

    The best response is the optimal strategy for player $i$ against the opponent strategy profile $\sigma_{-i}$, denoted as $b_i(\sigma_{-i})$. The value of the best response is $u_i(b_i(\sigma_{-i}), \sigma_{-i})$. Two-player zero-sum games have a game value, $v_i$, that is the lower bound on the utility of an optimal player in position $i$, formally,
    $$v_i = \min_{\sigma_i} u_i(\sigma_i, b_{-i}(\sigma_i)). $$
    In this case, the term \textbf{exploitability} of a strategy is how much additional utility is lost to the best response by playing $\sigma_i$, formally,
    $$ \varepsilon_i(\sigma_i) = v_i - u_i(\sigma_i, b_{-i}(\sigma_i)). $$

    In large two player zero-sum games the value of the game is unknown and is intractable to compute. However, if the players alternate positions, then the value of a pair of games is zero. If an agent players according to the profile $\sigma$ then its exploitability is
    \begin{equation}\label{eq:exploitability}
        \varepsilon(\sigma)=\frac{u_2(\sigma_1,b_2(\sigma_1))
        +u_1(b_1(\sigma_2),\sigma_2)}{2}
    \end{equation}

    \begin{figure*}[t]
    \centering
    \begin{subfigure}[b]{0.48\textwidth}
        \centering
        \includegraphics[width=\textwidth]{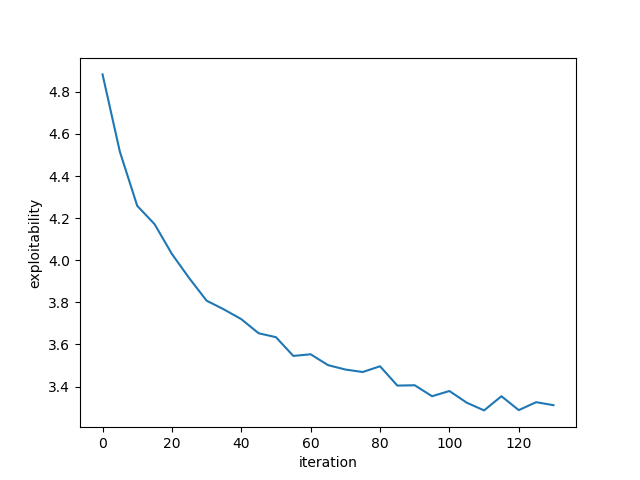}
        \caption{5 iterations in each epoch}
        \label{fig:expi5}
    \end{subfigure}
    \begin{subfigure}[b]{0.48\textwidth}
        \centering
        \includegraphics[width=\textwidth]{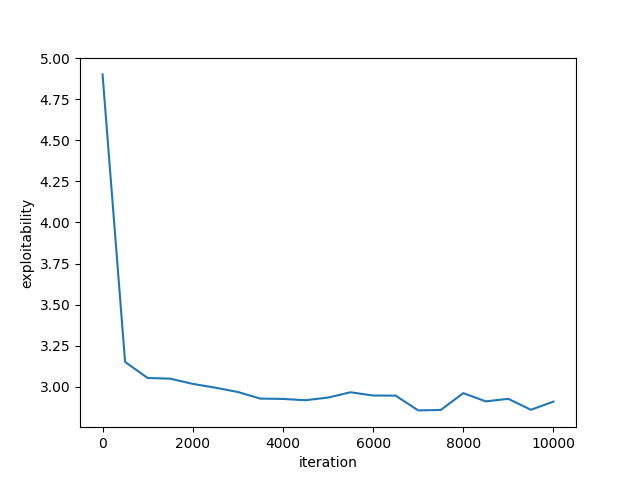}
        \caption{500 iterations in each epoch}
        \label{fig:expi500}
    \end{subfigure}
    \caption{Exploitability}
    \label{fig:exploitability}
\end{figure*}

    We run the experiments on a computer with Intel Core i9-9900 3.10GHz CPU and 16GM RAM. The training process is done with single thread and the evaluation is done with 100 threads. 3000 suits of tiles are randomly shuffled, serving as the benchmark for evaluation. For each of them, the CFR player plays with all heuristic search based players, and the opponent's highest score is selected as exploitability. Notice that this approximated exploitability is higher than the actual value, because it allows different actions on the same information set.

    The training process adopts CFR with Monte Carlo sampling. At every iteration, a suit of tiles is shuffled randomly and the CFR algorithm runs on this specific suit. The training process is done for a number of iterations in each epoch, and then it is evaluated on the benchmark generated previously.
    
    Such random training samples can hit the benchmark which shares the same feature encoding in Table~\ref{tab:encode}. Figure~\ref{fig:expi5} shows the training process where there are 5 iterations in each epoch. This experiment runs for 130 iterations and it takes around 12 hours. The exploitability gradually decreases but becomes slower after 100 iterations. Figure~\ref{fig:expi500} shows the case where each epoch has 500 iterations, which turns out to be long tail. The training and evaluating process takes around 72 hours. The exploitability drops significantly in the first 500 iterations but gradually becomes stable afterwards.

    After training for 10000 iterations, there are 2618 nodes in the database. Figure~\ref{tab:sqlite_example} shows an example of nodes. In round 1, there are 3 pairs and 2 Pongs or Kongs in hand, and 11 of out 14 tiles are wind tiles. Since there is no explicit Chows, Pongs or Kongs (All 14 tiles are in hand), all possible actions are legal, among which PongPongHu has the highest regret.

    \begin{table}[htbp]
        \centering
        \begin{tabular}{|c|c|}
        \hline
            encoding & 734401  \\ \hline
            round id & 1  \\ \hline
            number of Pairs & 3 \\ \hline 
            number of Pongs or Kongs & 2 \\ \hline 
            number of Character tiles & 3 \\ \hline 
            number of wind tiles & 11 \\ \hline 
            number of legal actions & 3 \\ \hline 
            regret sum & [-0.598, 2.128, -1.356] \\ \hline 
            strategy sum & [1.359, 1.975, 0.667] \\ \hline
        \end{tabular}
        \caption{An Example of Nodes}
        \label{tab:sqlite_example}
    \end{table}
    
\section{Discussion}\label{sec:discussion}
It's not entirely accurate to say that the hidden information stays the same during the entire game of Texas hold'em. As the game proceeds, additional cards are revealed, which reduces the number of possible states the players can be in. But this has only a minor impact on the number of states.

Games with private actions are an interesting problem and an area that has been under-explored. Algorithms like DeepCFR~\cite{brown2019deep} work without a problem in these games. Subgame solving~\cite{burch2014solving} works in theory, but subgame solving is only practical with a small amount of hidden information (less than about 1 million information sets per player per subgame). So subgame solving as it exists would not work in most games with private actions.

Compared to end-to-end algorithms, hierarchical policy abstraction is feasible and flexible. It takes comprehensive measures to deal with games with private actions, which can be trained on a single machine. Such algorithm can be extended to other Mahjong games with similar winning patterns. However, the current algorithm relies heavily on abstractions, which will be gradually released, such as the order of discarded tiles and the features of visible information.

\section{Acknowledgement}
This work was supported by Tencent Rhino-Bird Joint Research Program No. GF201911, in collaboration with Peng Sun at the Tencent AI lab.

The game logic is developed based on Sichuan Mahjong which was developed by Zhichao Shu from Tencent Lightspeed \& Quantum Studios Group. We add the logic of Chow and related winning patterns, formulate the game into extensive-form, and implement the recursive search framework.

We've received valuable advice from Noam Brown~\footnote{https://www.cs.cmu.edu/~noamb/} in regard to games with private actions and exploitability, which is discussed in section~\ref{sec:discussion}.

\bibliographystyle{aaai}
\bibliography{cfrgame}

\end{document}